\documentclass{article}

\usepackage{PRIMEarxiv}
\usepackage[utf8]{inputenc} 
\usepackage[T1]{fontenc}    
\usepackage{hyperref}       
\usepackage{url}            
\usepackage{booktabs}       
\usepackage{amsfonts}       
\usepackage{nicefrac}       
\usepackage{microtype}      
\usepackage{lipsum}
\usepackage{graphicx}
\usepackage[ruled,linesnumbered]{algorithm2e}
\usepackage{amssymb}
\usepackage{amsmath}
\usepackage{multirow}
\usepackage{booktabs}
\usepackage{subfigure}

\graphicspath{{media/}}     

\title{RNNCTPs: A Neural Symbolic Reasoning Method Using Dynamic Knowledge Partitioning Technology
}

\author{
	Yu-hao Wu \\
	School of Mathematical Sciences\\
	University of Electronic Science and Technology of China\\
	Chengdu, China \\
   \And
   Hou-biao Li \\
   School of Mathematical Sciences\\
   University of Electronic Science and Technology of China\\
   Chengdu, China \\
   \texttt{lihoubiao0189@163.com} \\
}

\begin{document}
\maketitle

\begin{abstract}
Although traditional symbolic reasoning methods are highly interpretable, their application in knowledge graph link prediction is limited due to their low computational efficiency. In this paper, we propose a new neural symbolic reasoning method: RNNCTPs, which improves computational efficiency by re-filtering the knowledge selection of Conditional Theorem Provers (CTPs), and is less sensitive to the embedding size parameter.
RNNCTPs are divided into relation selectors and predictors. The relation selectors are trained efficiently and interpretably, so that the whole model can dynamically generate knowledge for the inference of the predictor.
In all four datasets, the method shows competitive performance against traditional methods on the link prediction task, and can have higher applicability to the selection of datasets relative to CTPs.
\end{abstract}


\section{Introduction}
The concept of knowledge graph \cite{ehrlinger2016towards} proposed by Google in 2012 is a special knowledge base in which knowledge is composed of triples, which is developed from the semantic network.
The knowledge in the knowledge graph is composed of entities and relations, usually expressed in the form of triples of (entity, relation, entity), in which the entity is a concrete thing or abstract concept in reality, and the relation represents the relationship between entities associated information that exists.

Relationships in a triple are usually unidirectional, so a knowledge graph can be viewed as a directed graph connected by semantic relationships.
Due to the lack and incompleteness of the knowledge of the knowledge graph, it has caused great difficulties and restrictions on the use of the knowledge graph, and greatly restricted the retrieval, expansion and reasoning of related knowledge. Therefore, the need for knowledge map completion is imminent, and the knowledge map completion method emerges as the times require.

The mainstream knowledge graph completion methods are generally divided into two categories: the first category is knowledge extraction\cite{carlson2010toward}, which generally extracts new knowledge from unstructured text or image information; the second category is link prediction\cite{martinez2016survey}, use the existing knowledge in the knowledge graph to perform reasoning to complete the knowledge graph.
Due to the use of unstructured information, knowledge extraction requires a lot of computing resources, and the credibility of the extracted knowledge needs to be further verified, which is often difficult in specific use. The link prediction method performs reasoning in the structured information of the knowledge graph itself, which is more convenient to use.

Prevailing link prediction methods use neural models such as TransE\cite{bordes2013translating}, TransR\cite{lin2015learning}, RotatE\cite{sun2019rotate}, RESCAL\cite{nickel2011three}, DistMult\cite{yang2014embedding}, HOLE\cite{nickel2016holographic},
ComplEx\cite{trouillon2016complex}, ConvE\cite{dettmers2018convolutional}, which have wide applicability.
Due to the lack of interpretability of these models, it is not easy  to improve the performance of the models with the help of real-world experts. Link prediction methods also use symbolic reasoning models such as
AMIE\cite{galarraga2013amie,galarraga2015fast}, RLvLR\cite{omran2018scalable}, RuLES\cite{ho2018rule}, RPJE\cite{niu2020rule}, which faced with kinds of problems including weak generalisation results on datasets, discrete results which is unstable and hard to train with lots of modern optimization methods.

Neural symbolic reasoning methods such as NeuralLP \cite{yang2017differentiable}, MINERVA \cite{das2017go} have certain advantages in data robustness and reasoning process interpretability, but the challenges in computational efficiency make them difficult to generalize to large-scale knowledge graphs.
CILP++\cite{francca2015neural} focuses on logical induction or relational learning, which utilizes a rule extraction technique similar to that used by "model reduction", which can be thought of as being buildable by using logic as a constraint technique.
NTPs\cite{rocktaschel2017end,minervini2020differentiable} use the backward chaining algorithm to use the rules for reasoning, and can generate new rules. At the same time, it is robust to the use of reasoning predicates, but the operation speed is limited by the huge search space.
Kartasev\cite{kartavsev2021improving} uses reinforcement learning to improve the performance of the backward chain behavior tree, and also has a certain reference for the improvement path of the NTP method.
The two methods proposed by Xian improve the computational efficiency of similar methods by pre-capturing user behavior patterns\cite{xian2020neural} or rough sketches of user behavior\cite{xian2020cafe}, followed by fine-grained knowledge graph recommendation.
CTPs\cite{minervini2020learning} improves NTPs by adding a neural network layer for knowledge selection.
RNNNTPs\cite{wu2022neural} by leveraging RNN to capture the hierarchical structure of relations in the process of NTPs proof.
The reasoning process of CTPs and RNNNTPs is effectively pruned, which greatly improves the computational efficiency and can be used in large-scale knowledge graphs.

However, the sensitivity of CTPs to the embedding size parameter makes it stricter on the number of predicates in the dataset.
The inexplicability of knowledge selection of CTPs makes it difficult for CTPs to optimize with expert knowledge sometimes.
RNNNTPs will have a lot of redundant computation when a small number of predicates correspond to a large amount of knowledge.
We propose the RNNCTPs method, which can reduce parameter sensitivity, control the process of knowledge selection, and improve computational efficiency.
Next, we will briefly introduce NTPs and CTPs. In subsequent sections, we will introduce our model and conduct experiments.

\section{NTPs and CTPs}
NTPs\cite{rocktaschel2017end} and its conditional proving strategies optimised version CTPs\cite{minervini2020learning} are continuous
relaxation of the backward chaining algorithm: these algorithms works backward
from the goal, chaining through rules to find known facts supporting the proof.

Given a query(or goal) G, backward chaining first attempts to $unify$ it with the fact available in a given KB. If no matching fact is available, it considers all rules $H :- B$(We see facts as rules with no body and valuables), where $H$ denotes the head and $B$ the body, and $H$ can be unified with the query G resulting in a substitution for the variables contained in $H$. Then, the
backward chaining algorithm applies the substitution to the body $B$, and recursively attempts to prove until find the facts or catch the deep we have set.

Backward chaining can been seen as a type if $and/or$ search: $or$ means that any rule in the KB can be used to prove the goal, and $and$ means that all the premise of a rule must be proven recursively.

\textbf{Unification Module.}
In the backward chaining reasoning algorithm, $unification$\cite{rocktaschel2017end}
matches two logic atoms, such as $fatherOf(Anakin, Luke)$ and $dadOf(X, Y)$. It is backward chining reasoning's key operator, which play roles in discrete space.
In discrete spaces, equality between two atoms (e.g. $fatherOf \neq dadOf$ ) is evaluated by $unification$ by examining the elements that compose them, and using substitution sets
(e.g. ${X/Anakin, Y/Luke} $) binds variables to symbols. In NTPs, to be able to match different
symbols with similar semantics, $unify_\theta (H, G, S) = S' $ generate a neural network$-$a proof state
$S = (S_\psi , S_\rho )$ consisting of a set of substitutions $S_\psi$ and a proof score
$S_\rho$. For example, given a goal $G = [fatherOf, Anakin, Luke]$ and a fact $H = [dadOf, X, Y]$,the $unify$ module uses Gaussian kernel $k$ to compare the embedding representations of
$fatherOf$ and $dadOf$, updates the variable binding
substitution set $S_\psi'=S_\psi\cup\{{X/Anakin, Y/Luke}\}$, and calculates the new proof score $S_\rho'= min(S_\rho,k(\theta_fatherOf,\theta_dadOf))$ and proof state $S'=(S_\psi',S_\rho')$.

\textbf{OR Module.}
The $or$ module\cite{rocktaschel2017end} traverses a KB, computes the unification between goal and all facts and rule heads in it, and then recursively use the $and$ module on the corresponding rule
bodies. Given a goal $G$ and each rule $H :- B$ with the rule head $H$ in
a KB $\mathfrak{K} $, module $or_\theta^\mathfrak{K} (G, d, S)$ unifies the goal $G$ with the rule head $H$, and bodies $B$ of each rule will be proved by using module $and$ until reach the set deepest depth $d$
or $unify$ fail. $or$ module is shown below:
\begin{equation}
    \begin{split}
        or_\theta^\mathfrak{K} (G, d, S) &= [S'|H:-B \in \mathfrak{K} ,\\
        &S' \in and_\theta^\mathfrak{K} (B, d, unify_\theta(h, g, s))].
    \end{split}
\end{equation}

\textbf{AND Module.}
After unification in $or$ module\cite{rocktaschel2017end}, $and$ module proves a list of sub-goals in a rule body $B$.
$and_theta^\mathfrak{k}(B:\mathbb{B},d,S)$ module first substitute variables in the first sub-goal B with constants
using substitutions in $S$, then use $or$ module to generate another sub-goals of B. $\mathbb{B}$ use the result state of above to prove the atoms by using the $and$ module recursively:
\begin{equation}
    \begin{split}
        and_\theta^\mathfrak{K} (B : \mathcal{B} , d, S) = &[S'' | d > 0, \\
        &S'' \in and_\theta^\mathfrak{K} (\mathcal{B} , d, S'), \\
        &S' \in or_\theta^\mathfrak{K} (sub(B, S_\phi), d - 1, S)].
    \end{split}
\end{equation}

\textbf{Proof Aggregation.}
In a KB $\mathfrak{k}$, we use modules above to generate a neural network, which evaluates all the
possible proofs of a goal $G$. The largest proof score will be selected by NTPs:
\begin{equation}
    \begin{split}
        ntp_\theta^\mathfrak{K} (G, d)=max_SS_\rho \\
        with S \in or_\theta^\mathfrak{K} (G,d,(\emptyset,1)).
    \end{split}
\end{equation}
where $d \in \mathbb{N}$, and $\mathbb{N}$ is defined at the beginning.
The initial proof state is set to $(\emptyset,1)$ which express an empty
substitution set and a proof score of 1.

\textbf{Training.}
NTPs minimise a cross-entropy loss $\mathcal{L}^\mathfrak{K} (\theta)$
to learn embedding representations of atoms using the final proof score.
Prover masking facts iteratively and try to prove them using other facts and rules.

CTPs is an upgraded version of NTPs. By adding a neural layer for selecting knowledge, CTPs reduces the amount of knowledge to be traversed in each traverse, so as to achieve the effect of improving computing efficiency.The CTPs $OR$ module has been rewritten as follows:

\begin{algorithm}[H]
  \caption{Or Module in CTPs\cite{minervini2020learning}.}
  \label{al or ctps}
  \textbf{function} or(G, d, S):\\
  \For{$H :- \mathbb{B} \in select_\theta(G)$}{
      \For{$S\in and(\mathbb{B},d,unify(H,G,S))$}{
          yield S
      }
}
\end{algorithm}

Algorithm \ref{al or ctps} shows how CTPs divides the knowledge required for this traversal according to the input proof goal $G$. The key is the selection function $select_\theta(G)$, which is introduced in CTPs. Different selection functions are selected according to different data sets to achieve better calculation results.

For example, linear function-based selection modules are used in UMLS\cite{bodenreider2004unified} and Nations\cite{kok2007statistical}:
\begin{equation}
  select_{\theta}(G) \triangleq \left[F_{\mathrm{H}}(G):-F_{\mathrm{B}_{1}}(G), F_{\mathrm{B}_{2}}(G)\right],
\end{equation}
knowledge of the rules of the head and body separation by $F_H(G)=[f_H(\theta_{G_1}),X,Y]$, $F_{B_1}(G)=[f_{B_1}(\theta_{G_1}),X,Z]$ and $F_{B_2}(G)=[f_{B_2}(\theta_{G_1}),Z,Y]$.
$f_i : \mathbb{R}^k\rightarrow \mathbb{R}^k$ is a differentiable function that can be trained end-to-end, similar to a linear function $f_i(x)=W_i x + b$,
where $W_i \in \mathbb{R}^{k\times k}$ and $b\in \mathbb{R}^k$.

The Kinship\cite{jenatton2012latent} data set uses a selection module based on the attention function, which is as follows:
\begin{equation}
  \begin{aligned}
  f_{i}(\mathbf{x}) &=\alpha \mathbf{E}_{\mathcal{R}} \\
  \alpha &=\operatorname{softmax}\left(\mathbf{W}_{i} \mathbf{x}\right) \in \Delta^{|\mathcal{R}|-1},
  \end{aligned}
\end{equation}
where $\mathbf{E}_{\mathcal{R}}\in \mathbb{R}^{\mathcal{R}\times k}$ represents the embedding matrix of the predicate, $\mathbf{W}_i\in \mathbb{R}^{k\times \mathcal{R}}$, $\alpha$ is the attention distribution of a predicate on $\mathcal{R}$. When such attention-based selection function is used, it is generally default to set a reasonable embedding dimension greater than the number of predicates in the data set and model, so that the effect of attention mechanism can be better played. CTPs also uses the mechanism of memory network\cite{miller2016key} to strengthen the selection module and solve the problem of redirection of rule selection.

\section{RNNCTPs Algorithm Based on Dynamic Knowledge Region Division}
\begin{figure}[htbp]
  \centering
  \includegraphics[scale = 0.6]{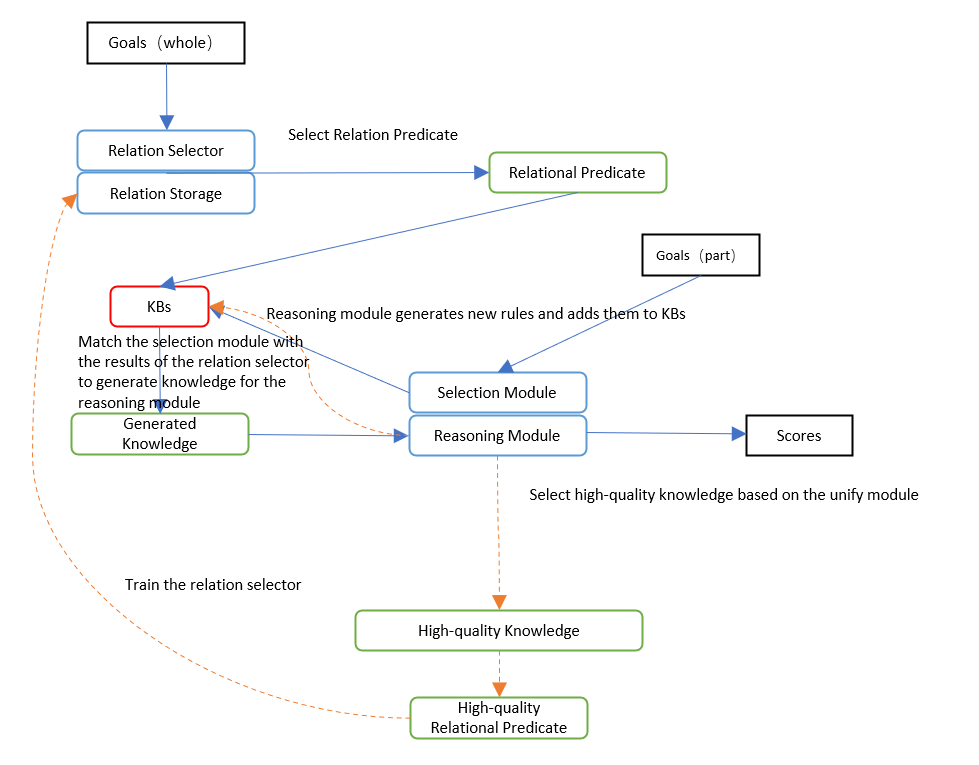}
  \caption{RNNCTPs Model Framework. Inputting a batch of relational predicates proving goals, the relational selector generates a set of relational predicates. In the specific proof stage, each knowledge generated by goal is matched as shown in Figure \ref*{fig:match}, and finally a series of knowledge is dynamically generated for reasoning. During the inference phase, high-quality knowledge is also returned for the training of relation selectors.}
  \label{fig:rnnctps}
\end{figure}

In order to take advantage of the dynamic knowledge selection feature of CTPs,
we proposes a link prediction algorithm based on dynamic knowledge region generation called RNNCTPs, see Figure \ref*{fig:rnnctps}.
The CTPs module uses neural network for dynamic knowledge selection, and RNN-based relationship generator generates a series of relationship sequences,
and then matches the knowledge dynamically selected by CTPs according to the relationship contained in the knowledge.
Similarly, the relationship generator is trained by marking the knowledge used after the CTPs proof is completed.

\subsection{Relation Generator}
The reasoning part of CTPs proves that the process is hierarchical, and sequence relations exist in the knowledge utilized by different layers.
We use RNN as the generator of relational predicates.
The knowledge generated by the CTPs selection module will be selected again,
so we set a range expansion in the initial knowledge generation to limit the amount of knowledge ultimately matched and used in the reasoning part of the proof process to a reasonable range.
The scope of knowledge selection depends on the number of input batch proof targets, and the number of relationships generated by the relation generator also depends on the success rate of knowledge matching in CTPs. In the case that the correlation and hierarchical structure of generating relationships can be fully reflected,
generating a reasonable number of relationships and making the number of successful matching knowledge in a relatively stable and controlled range is the problem that our relationship generator needs to solve.
The rest of this chapter calls the relationship generator $RNN_\theta$.

\textbf{Relation Selector. }The rule structure is $H:-B$, where $H$ is the rule head, $B$ is the rule body, and rule body $B$ consists of sequence relations $r_1,r_2,\dots,r_n$. We uses a relation selector $RNN_\theta$ to predict the sequence relation. Given a batch of prediction target set $G$, take out its relation $R$ and initialize $RNN_\theta$:
\begin{equation}
	h_0=f(r),
\end{equation}
then use GRU gating unit to get hidden variables at different positions:
\begin{equation}
	h_t=GRU(h_{t-1},g([r_{t-1},r_t])),
\end{equation}
where $G$ is a linear transformation, $[r,r_t]$ is a connection between the previous rule body (rule head) and the current rule body.

By using $softmax(o(h_{t+1}))$, a probability partition is carried out on the generated rule set $R$, and the next rule body $r_{t+1}$ is obtained.Each rule body $r_i$ is added to a relational set $Rel_{gen}$.

The control of $relNum$, $KB_\lambda$ of the number of generated relations becomes relatively dynamic here, which is obtained by the following formula:
\begin{equation}
  relNum = batchSize\times size(Rel_{storage})\times rel_\lambda \times KB_\lambda,
\end{equation}
where $batchSize$ is the target number of batch proof, $size(Rel_{storage})$ is the number of relationships in the relational warehouse, $relNum_\lambda$ is the hyperparameter to adjust the number of relationships generated, $KB_\lambda$ is the matching degree of the knowledge generated by CTPs and the relationship generated by the relationship generator,
which will be set in detail in the introduction of the predictor. In practice,
the number of relationships generated by each proof target is actually used to limit the number of relationships generated:
\begin{equation}
  relNumPer = size(Rel_{storage})\times relNum_\lambda \times KB_\lambda.
\end{equation}

\textbf{Relation Storage.}
Because the reasoning module of the predictor makes use of dynamically generated knowledge domain,
we proposes a relational storage structure based on the previous section, in order to train the relational generator better by using the return information of the reasoning module.
Use a set of dynamic extension coefficients $ \{ep_1, ep_2,\dots, ep_{maxDepth} \}$, related to the depth of the relationship in the proof process, the storage structure is as follows:
\begin{equation}
	Rel_{storage} = \{ \overbrace{r_1, \cdots, r_j}^{ep_1}, \overbrace{r_{j+1}, \cdots, r_{k}}^{ep_2*(j-1+1)}, \overbrace{r_{k+1}, \cdots, r_l}^{ep_3*(k-j)}, \cdots \}.
\end{equation}
Expansion coefficients $ \{ep_1, ep_2,\dots, ep_{maxDepth} \}$ dynamically change as follows:
\begin{equation}
	ep_i^{new} = ep_i^{last} - \kappa (KB_\lambda^{new} - KB_\lambda^{last}),
\end{equation}
where $\kappa$ is the hyperparameter controlling $ep$ adjustment and is a positive real number. $KB_\lambda^{new}$ is the current $KB_\lambda$, $KB_\lambda^{last}$ is the last period $KB_\lambda$.
When the degree of knowledge matching is higher, relatively fewer relationships can be added to the relationship warehouse, and the current hidden distribution can be fine-tuned.

\begin{figure}[htbp]
	\centering
	\includegraphics[scale = 0.6]{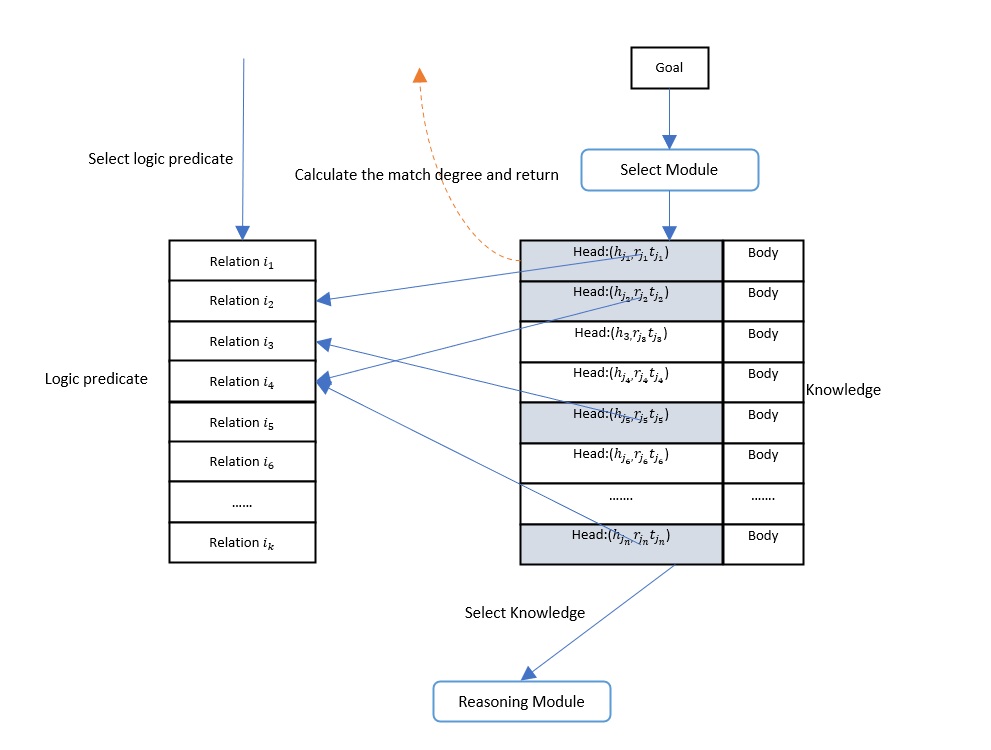}
	\caption{Candidate Knowledge Matching Process. Each goal will generate a series of knowledge through the selection layer of CTPs, and then filter through the relationship predicate generated by the relationship selector, and input the filtered knowledge into the inference module for inference.}
	\label{fig:match}
\end{figure}

\subsection{CTPs Based Predictor}

The improvement of CTPs in our model mainly lies in redirecting the knowledge obtained after module selection with memory network, obtaining a new knowledge set through matching and filtering with the relation set generated by the relation generator, and then using it in the inference module to form a new sub-proof goals set. The relationship generation returns the success rate of knowledge matching in CTPs, so as to adjust the number of relations generated in the relation generator.

As shown in Figure \ref{fig:match}, CTPs generates a series of knowledge, including rule head and rule body, according to the proof objective through the selection module.
According to the above mentioned memory network, RNNCTPs method can extract the corresponding relational predicates from the embedded matrix and match with the predicates generated by the relational generator according to the batch proof object set.
If the relational predicate in the knowledge exists in the predicate generated by the relational generator, it means that the knowledge is successfully matched and can be used successfully and sent into the reasoning module of CTPs for the following operations.
Note that the CTPs selection module is called at each level of the proof (the subproof object replaced by the rule body is used again by the selection module), whereas the relationship generator only generates the relationship once after a batch of proof object input.

After CTPs generates knowledge set each time, when matching is completed, variable set \{$KB_{\lambda_i}$\} needs to be recorded:
\begin{equation}
  KB_{\lambda_i}=\frac{\lvert KB_{Used_i} \rvert}{\lvert KB_{gen_i} \rvert},
\end{equation}
$KB_{Used_i}$ represents the knowledge used this time, and $KB_{gen_i}$ represents the knowledge generated by CTPs this time.
Calculate the $KB_{\lambda_{mean}}$ return of \{$KB_{\lambda_i}$\} for training of relational selector.

\subsection{Algorithm Process}
The approximate EM algorithm is  used to optimize the dynamic selection of rules for the whole model. RNN will be set to make the size of the knowledge set given to the reasoning module of CTPs to reason within a reasonable range.
Therefore, in RNNCTPs, the number of RNN-generated relational predicates needs to be adjusted according to the success degree of matching between RNN-generated relational predicates and the knowledge set generated by CTPs selection module, so that the number of generated relational predicates reaches a dynamic balance.

\begin{algorithm}[htbp]
	\caption{Pseudocode for CTPs Algorithm Training Based on Dynamic Knowledge Partitioning.}
	\label{alg:algorithm2}
	\KwIn{triple data set goals$[(h,r,t)]$, number of iterations $n$.}
	\KwOut{Scores, trained generator $RNN_\theta$, trained predictor $CTPs_\theta$.}
	\BlankLine
	Initialize $KB$, $RNN_\theta$, $CTPs_\theta$;
	
	\For{i = 1 to n}{
		$RNN_\theta$ generate $Rel_{gen}$\;
		\textbf{e-step:} start train $CTPs_\theta$:\\
		$KB_{selected}$ = $Select_\theta(G)$\;
        $KB_{gen} = match(KB_{selected}, Rel_{gen})$\;
		\If{$unify(facts\ or\ rules\in KB_{gen})$ not FAIL}{
			$KB_{high-quality}$ add$(facts\ or\ rules)$;
		}
		\While{size$(Rel_{storage})$ < maxSize$(Rel_{storage})$}{
			$KB_{high-quality}$ add $NNS(KBs)$\;
			$Rel_{storage}$ add predicates of $KB_{high-quality}$\;
		}
		Relation Selector add  \{$KB_{\lambda_i}$\}\;
		\textbf{m-step:} train Relation Selector with $Rel_{storage}$\;
		train Relation Selector with \{$KB_{\lambda_i}$\};
	}
\end{algorithm}

Algorithm \ref{alg:algorithm2} shows the main pseudo-code flow of CTPs algorithm based on dynamic knowledge partition.
In RNNCTPs, the process of knowledge generation is changed to a match after the CTPs uses the knowledge according to the proof objective generation: firstly, $Select_\theta(G)$ in CTPs is called to generate $KB_{selected}$.
Then the relational predicates in $Rel_{gen}$ and $KB_{selected}$ are matched to dynamically generate the knowledge set $KB_{gen}$ needed to prove the current target. When entering the next subobject of proof, the knowledge set required by the subobject is dynamically generated until the set proof depth is reached.
The training of CTPs has already included the training of its own selection modules.
The number of generated relation predicates needs to be dynamically adjusted by the relationship selector to make the total amount of knowledge used in the calculation of the $unity$ module in a controllable range.
In the e-step, our predictor uses the knowledge in this domain to score this batch of goals, and put the useful knowledge into the $High-quality knowledge$.
When the size of $ Relation Storage$ does not reach the set value,
we need to use Nearest Neighbor Search(NNS)\cite{tao2002continuous} to search for $High-quality Knowledge$ in KB first, and add the predicate(relation) of $High-quality Knowledge$ to $Relation Storage$.

In the m-step, we train relation selector $RNN_\theta$ : the knowledge utilization set \{$KB_{\lambda_i}$\} is generated after each iteration, and the mean $KB_{\lambda_{mean}}$ is calculated.
$RNN_\theta$ adjustment $KB_\lambda$ by $KB_{\lambda_{mean}}$:
\begin{equation}
  KB_\lambda = \begin{cases}
      KB_\lambda+\tau (KB_{\lambda_{mean}}-KB_\lambda), &if \lvert KB_{\lambda_{mean}}-KB_\lambda \rvert > \varsigma\\
      \frac{KB_\lambda+KB_{\lambda_{mean}}}{2} , &else	
             \end{cases}
\end{equation}
$\tau$ is the updated learning rate of $KB_\lambda$ and $\varsigma$ is the learning threshold.In order to prevent $KB_\lambda$ from being influenced too much by a small number of partial matches or knowledge sets that are too small or too large, only controllable fine-tuning is performed when the batch $KB_{\lambda_{mean}}$ differs too much from $KB_\lambda$.

\section{Experiment}
\subsection{Experimental Setup}
\textbf{Datasets: Kinship, UMLS, Nations and Countries\cite{bouchard2015approximate}.} In the experiments, we selected four benchmark datasets, namely Alyawarra Kinship (Kinship), Unified Medical Language System (UMLS), Nations. Nations contains 56 binary predicates, 111 unary predicates, 14 constants and 2565 true facts; Kinship contains 26 predicates, 104 constants and 10686 true facts; UMLS contains 49 predicates, 135 constants and 6529 true facts real facts. Because our benchmark ComplEx cannot handle unary predicates, we remove unary atoms from Nations. For Kinship, UMLS, and Natons datasets, there is no standard data split. So we split each knowledge base into 30\% training facts, 20\% validation facts and 50\% test facts. For evaluation, we take a test fact and corrupt its first and second arguments in all possible ways so that the corrupted fact is not in the original KB. Subsequently, we predict each test fact and its damage level, and compute MRR and HITS@m. We also used a country dataset\cite{trouillon2016complex} to evaluate the scalability of our algorithm. This dataset contains 272 constants, 2 predicates, and 1158 ground truths, and is designed to explicitly test the logical rule induction and reasoning abilities of link prediction models.

\textbf{Evaluation benchmark. }In Kinship, UMLS, Nations and Countries, we predict each test fact and its associated rank, and compute MRR, HITS@m(m=1,3,10) after training. In countries, we evaluate the area under the precision-recall curve (AUC-PR) and the results are compared with previous methods. Set the average training time per iteration (relative) (ATTP) to compare computational performance. Knowledge utilization refers to each time we use knowledge, and it is measured by the successful establishment of the unified branch and comparing the target utilization of this batch of knowledge. ATTP and Knowledge Utilization can evaluate expected performance on large datasets.

CTPs and RNNCTPs using linear functions and attentional selection modules as well as other classical methods need to be tested on different data sets.
Then the focus is on the comparison of RNNNTPs\cite{wu2022neural} using attentional selection module, and the comparison of accuracy and performance indicators between CTPs and the set of relations of different embedding dimensions and sizes.

The ratio for $KB_{gen}$ is set to 30\%.The number of storage tiers of $Relation Storage$ is the same as the number of CTPs recursive tiers and set it to 3. The number of predictor iterations was set to 100, and every 20 iterations, the relationship generator was trained.

\subsection{Experimental Results}
\subsubsection{Link Prediction Task Evaluation}

\begin{table}[htbp]
	\centering
	\caption{Experimental Accuracy Evaluation of RNNCTPs on Link Prediction.}
	\label{table RNNCTPs}
	\begin{tabular}{lllllllll}
	\toprule
	\multirow{2}{*}{Datasets}  &    & \multirow{2}{*}{Metrics} & \multicolumn{6}{c}{Models}                                                                         \\
	\cline{4-9}
							   &    &                          & RNNCTPs & RNNNTPs & CTPs  & NTPs                                             & NeuralLP & MINERVA  \\
	\hline
	\multirow{4}{*}{Nations}   &    & MRR                      & 0.708   & 0.701   & 0.709 & 0.74                                             &          &          \\
							   &    & HITS@1                   & 0.583   & 0.539   & 0.562 & 0.59                                             &          &          \\
							   &    & HITS@3                   & 0.827   & 0.875   & 0.813 & 0.89                                             &          &          \\
							   &    & HITS@10                  & 0.998   & 0.997   & 0.995 & 0.99                                             &          &          \\
	\multirow{4}{*}{Kinship}   &    & MRR                      & 0.742   & 0.772   & 0.764 & 0.80                                             & 0.619    & 0.720    \\
							   &    & HITS@1                   & 0.632   & 0.609   & 0.646 & 0.76                                             & 0.475    & 0.605    \\
							   &    & HITS@3                   & 0.834   & 0.891   & 0.859 & 0.82                                             & 0.707    & 0.812    \\
							   &    & HITS@10                  & 0.943   & 0.973   & 0.958 & 0.89                                             & 0.912    & 0.924    \\
	\multirow{4}{*}{UMLS}      &    & MRR                      & 0.812   & 0.801   & 0.852 & 0.93                                             & 0.778    & 0.825    \\
							   &    & HITS@1                   & 0.621   & 0.589   & 0.752 & 0.87                                             & 0.643    & 0.728    \\
							   &    & HITS@3                   & 0.934   & 0.951   & 0.947 & 0.98                                             & 0.869    & 0.900    \\
							   &    & HITS@10                  & 0.974   & 0.972   & 0.984 & 1.00                                             & 0.962    & 0.968    \\
	\multirow{3}{*}{Countries} & S1 & \multirow{3}{*}{AUC-PR}  & 100.0   & 100.0   & 100.0 & \begin{tabular}[c]{@{}l@{}}100.00\\\end{tabular} & 100.00   & 100.0    \\
							   & S2 &                          & 91.28   & 89.47   & 91.81 & 93.04                                            & 75.1     & 92.36    \\
							   & S3 &                          & 93.22   & 95.21   & 94.78 & 77.26                                            & 92.20    & 95.10    \\
	\bottomrule
	\end{tabular}
\end{table}

Table \ref{table RNNCTPs} shows the results of link prediction experiment. The best parameters of RNNCTPs and CTPs in each data set and the situation of selecting modules are selected respectively. It can be seen that in the Nations dataset and the three evaluation indicators, RNNCTPs are superior to CTPs. This is because a rule head in the Nations data set is often accompanied by a rule body with two elements. In some cases, the knowledge generated by CTPs has too many proof paths, so it cannot be targeted at training.
And RNNCTPs is filtered, so that the selection module of CTPs can also be trained according to the feedback of the proof process.
The results were in line with expectations. In other datasets, because of the reduction of computation and the rule head is often accompanied by the rule body containing 1 element, the proof path of interference is relatively small, CTPs can carry out perfect training, so RNNCTPs has a gap with CTPs and other leading methods and classical methods.

In Table \ref{table Kinship RNNCTPs}, RNNCTPs showed a stronger experimental effect than CTPs with linear selection module. The small number of relationship predicates in Kinship is conducive to the play of the attention selection module, while RNNCTPs also filters the knowledge generated by the linear selection module.
Therefore, the experimental effect of RNNCTPs with linear selection module and attentional selection module is significantly better than that of CTPs with linear selection module, and the RNNCTPs with attentional selection module is only slightly lower than that of CTPs with attention module. Experiments on the Kinship dataset are also expected.

\begin{table}[htbp]
  \centering
  \caption{Evaluation of Link Prediction Task on Kinship Dataset.}
  \label{table Kinship RNNCTPs}
  \begin{tabular}{lllll}
  \toprule
  Model              & MRR   & HITS@1 & HITS@3 & HITS@10  \\
  \hline
  RNNCTPs(Linear)    & 0.677 & 0.570  & 0.842  & 0.948    \\
  RNNCTPs(Attentive) & 0.742 & 0.632  & 0.834  & 0.943    \\
  CTPs(Linear)       & 0.655 & 0.457  & 0.756  & 0.891    \\
  CTPs(Attentive)    & 0.764 & 0.646  & 0.859  & 0.958    \\
  \bottomrule
  \end{tabular}
  \end{table}

  \begin{table}[htbp]
    \centering
    \caption{Evaluation of Link Prediction Task on UMLS Dataset.}
    \label{table UMLS RNNCTPs}
    \begin{tabular}{lllll}
    \toprule
    Model              & MRR   & HITS@1 & HITS@3 & HITS@10  \\
    \hline
    RNNCTPs(Linear)    & 0.762 & 0.599  & 0.915  & 0.923    \\
    RNNCTPs(Attentive) & 0.812 & 0.621  & 0.934  & 0.974    \\
    CTPs(Linear)       & 0.852 & 0.752  & 0.947  & 0.984    \\
    CTPs(Attentive)    & 0.732 & 0.685  & 0.912  & 0.901    \\
    \bottomrule
    \end{tabular}
\end{table}

Table \ref{table UMLS RNNCTPs} shows the excellent performance of RNNCTPs in the multi-relational predicate data set. Since the number of relationships generated by the relationship generator in RNNCTPs according to the current proof target batch each time is smaller than the dimension of the embedding matrix of CTPs attention selection module, the attention selection module can be used to improve the effect of the model. It can be seen that CTPs using attentional selection module has inferior effect to CTPs using linear selection module, while RNNCTPs using attentional module is not only better than CTPs using attentional module, but also better than CTPs using linear selection module.

\subsubsection{Computational Performance}

\begin{figure}[htb]
	\centering
	\subfigure[]{
		\includegraphics[height=5.57cm,width=7cm]{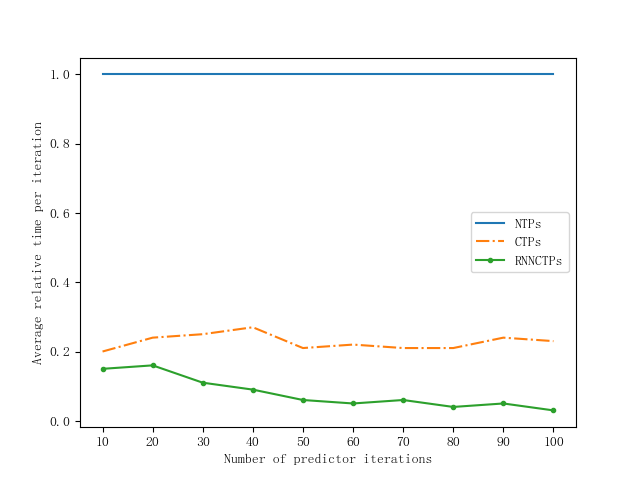}		
	}
	\subfigure[]{
		\includegraphics[height=5.57cm,width=7cm]{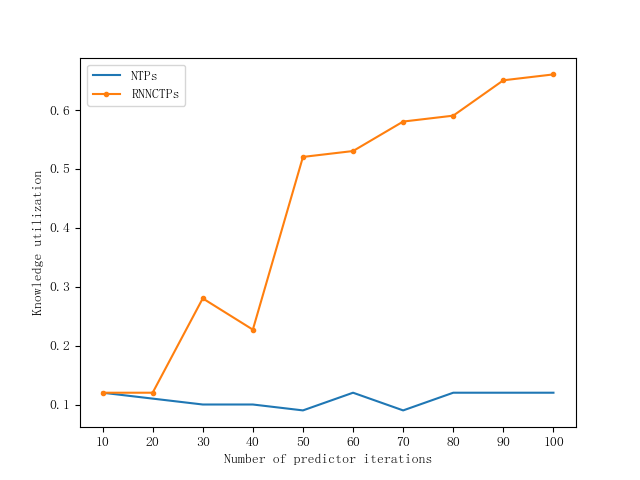}
	}
	\caption{Overall performance of RNNCTPs. As the number of iterations increases, the trained relation generator can more effectively combine the selection layer of CTPs to select knowledge, and the overall computing performance is improved.}
	\label{ctp compute}
\end{figure}

As shown in figure \ref{ctp compute}, the overall performance of RNNCTPs is significantly improved compared with that of NTPs.
This is due to the re-selection of the knowledge of CTPs selection module, and the performance has been improved compared with CTPs.
In terms of knowledge utilization, after every 20 iterations of relation generator training, compared with NTPs, it also has a great improvement. It can be seen that the overall performance is excellent, especially compared with NTPs.
In terms of both iteration time and knowledge utilization, great progress has been made to meet the pre-target.

\begin{figure}[htb]
	\centering
	\subfigure[]{
		\includegraphics[height=5.57cm,width=7cm]{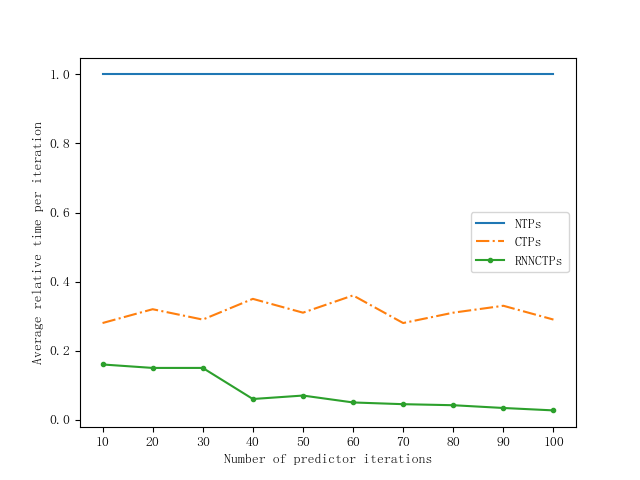}		
	}
	\subfigure[]{
		\includegraphics[height=5.57cm,width=7cm]{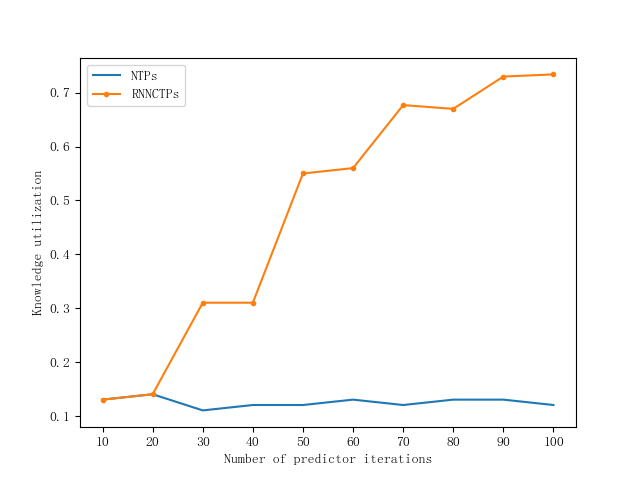}
	}
	\caption{Computational performance of RNNCTPs on Nations dataset. On the Nations dataset with more complex rules, RNNCTPs can better improve the computational efficiency.}
	\label{ctp nations compute}
\end{figure}

On the Nations data set, RNNCTPs achieved better performance on tasks with relatively complex multi-element rule bodies.As shown in figure \ref{ctp nations compute}, RNNCTPs performed better in the case of more proved path forking compared to the overall performance.
RNNCTPs has the ability to complete knowledge graph under more complex rules.

\begin{figure}[htb]
	\centering
	\subfigure[]{
		\includegraphics[height=5.57cm,width=7cm]{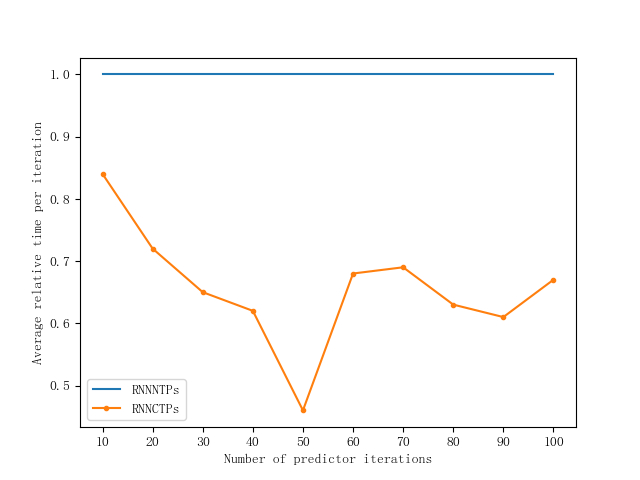}		
	}
	\subfigure[]{
		\includegraphics[height=5.57cm,width=7cm]{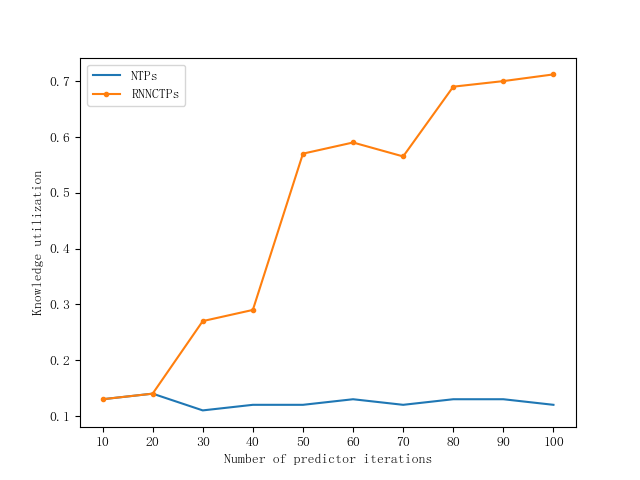}
	}
	\caption{Computational performance of RNNCTPs on Kinship dataset. The Kinship dataset has more knowledge relative to relational predicates, and RNNCTPs also show the computational efficiency improved by the matching filtering system.}
	\label{ctp kinship compute}
\end{figure}

Some relational predicates in the Kinship dataset are associated with far more knowledge than other relational predicates, and the overall number of predicates is relatively small. As shown in figure \ref{ctp kinship compute}, RNNCTPs have a relatively obvious improvement in performance compared to RNNNTPs. This is because RNNNTPs only generate knowledge based on relational predicates, while RNNCTPs will perform knowledge based on the proof target. An option that narrows the search considerably. It can also be seen that RNNCTPs also have good performance for datasets with a large amount of knowledge associated with relational predicates. Although the selection module in the CTPs module consumes part of the performance, on such datasets, RNNCTPs can reduce the number of knowledge traversals. to get performance improvement. In summary, RNNCTPs have the ability to complete knowledge graphs under multi-entity and multi-simple relationships.

\subsubsection{Parameter Sensitivity}
Mainly compare the Kinship dataset with 26 predicates and 10696 factual knowledge and the UMLS dataset with 49 predicates and 6529 factual knowledge in the \textbf{embedding size parameter} of the inference module based on the linear selection module and attention, respectively. Comprehensive performance of CTPs on the selection module and RNNCTPs based on the attention selection module.

\begin{table}[htbp]
	\centering
	\caption{Evaluation of Prediction Tasks with Different Embedding Sizes on Kinship Dataset.}
	\label{table Kinship Embedding}
	\begin{tabular}{llllllllll}
	\toprule
	Model          & \multicolumn{3}{l}{RNNCTPs(Attentive)} & \multicolumn{3}{l}{CTPs(Attentive)} & \multicolumn{3}{l}{CTPs(Linear)}  \\
	\cline{2-10}
	Embedding Size & 30    & 40    & 50                     & 30    & 40    & 50                  & 30    & 40    & 50                \\
	\hline
	MRR            & 0.741 & 0.738 & 0.742                  & 0.598 & 0.712 & 0.764               & 0.654 & 0.678 & 0.697             \\
	HIT@1          & 0.612 & 0.623 & 0.632                  & 0.501 & 0.621 & 0.646               & 0.512 & 0.554 & 0.587             \\
	HIT@3          & 0.849 & 0.854 & 0.834                  & 0.762 & 0.867 & 0.859               & 0.788 & 0.754 & 0.821             \\
	HIT@10         & 0.941 & 0.935 & 0.943                  & 0.832 & 0.943 & 0.973               & 0.861 & 0.873 & 0.938             \\
	\bottomrule
	\end{tabular}
\end{table}

Table \ref{table Kinship Embedding} shows the performance of the link prediction tasks of the three models under the Kinship data set under the embedding sizes of different inference modules.
The optimal case of CTPs is to use the selection module with attention when the embedding size is 50, but it has a significant decline when the embedding size is 40, and a huge decline when the embedding size is 30.
This is because there are 26 predicates in the Kinship data set, and the attention mechanism alone is weak when the embed size is 30, and the amount of knowledge generated is less than RNNCTPs, so many predicates that can be used remain unselected.
RNNCTPs is not so sensitive to parameters.
This is because RNNCTPs selects knowledge together with the relation generator and the attention module, and the amount of knowledge generated by the attention module is large.
After filtering by the predicate generated by the relation generator, more useful information enters the next layer of the proof path, which meet design expectations.

\begin{table}[htbp]
	\centering
	\caption{Evaluation of Prediction Tasks with Different Embedding Sizes on the UMLS Dataset.}
	\label{table UMLS Embedding}
	\begin{tabular}{llllllllll}
	\toprule
	Model          & \multicolumn{3}{l}{RNNCTPs(Attentive)} & \multicolumn{3}{l}{CTPs(Attentive)} & \multicolumn{3}{l}{CTPs(Linear)}  \\
	\cline{2-10}
	Embedding Size & 30    & 40    & 50                     & 30    & 40    & 50                  & 30    & 40    & 50                \\
	\hline
	MRR            & 0.807 & 0.804 & 0.812                  & 0.545 & 0.675 & 0.712               & 0.712 & 0.827 & 0.852             \\
	HIT@1          & 0.616 & 0.622 & 0.621                  & 0.512 & 0.637 & 0.601               & 0.692 & 0.741 & 0.752             \\
	HIT@3          & 0.874 & 0.912 & 0.934                  & 0.748 & 0.845 & 0.868               & 0.817 & 0.891 & 0.947             \\
	HIT@10         & 0.936 & 0.956 & 0.974                  & 0.816 & 0.897 & 0.904               & 0.869 & 0.901 & 0.984             \\
	\bottomrule
	\end{tabular}
\end{table}

Table \ref{table UMLS Embedding} shows the performance of the link prediction task of the three models under the UMLS data set under the embedment sizes of different reasoning modules. CTPs has the best effect when the linear selection module is used and the embedding size is 50. Its attention module has a low effect and is basically unavailable.
This is because when UMLS contains 49 predicates and the embedding size is 50, the effectiveness of the attention mechanism to select predicates has been significantly reduced. RNNCTPs selects knowledge together with the relation generator and the attention module, and has strong fault tolerance for the knowledge selected from the attention layer, which is slightly better than the linear layer RNNCTPs after testing in the actual state.

After the experiment, RNNCTPs is less dependent on the parameters of the embedding layer size, and the result gap is not obvious after the embedding layer size. The experimental effect gap is not significant when the embedding layer size is reduced by 20\%, and the good result can be achieved when the embedding layer parameters are reduced by 40\%.

\section{Conclusions}
We propose a neural symbolic reasoning method based on dynamic knowledge search region division: RNNCTPs, and introduce the relation generation module, inference prediction module and connection matching module of RNNCTPs in detail.
In view of the inadequacy of CTPs in knowledge selection and the sensitivity of its attention selection module to embedding size, RNNCTPs solves the above problems in link prediction task.
In the subsequent experiment, in the open datasets, RNNCTPs has further improved the computational efficiency compared with previous models, and is less sensitive to embedding size parameters in the Kinship dataset.
In UMLS, attention selection module can be used to improve the overall effect even when the number of predicates is larger than the embedding size.
RNNCTPs can still be interpreted in relation selection, and this module can conduct additional rule training to improve the overall training effect.


\bibliographystyle{unsrt}
\bibliography{references}

\begin{thebibliography}{10}

\bibitem{ehrlinger2016towards}
Lisa Ehrlinger and Wolfram W{\"o}{\ss}.
\newblock Towards a definition of knowledge graphs.
\newblock {\em SEMANTiCS (Posters, Demos, SuCCESS)}, 48(2):1--4, 2016.

\bibitem{carlson2010toward}
Andrew Carlson, Justin Betteridge, Bryan Kisiel, Burr Settles, Estevam~R
  Hruschka, and Tom~M Mitchell.
\newblock Toward an architecture for never-ending language learning.
\newblock In {\em Twenty-Fourth AAAI conference on artificial intelligence},
  pages 1306--1313, 2010.

\bibitem{martinez2016survey}
V{\'\i}ctor Mart{\'\i}nez, Fernando Berzal, and Juan-Carlos Cubero.
\newblock A survey of link prediction in complex networks.
\newblock {\em ACM computing surveys (CSUR)}, 49(4):1--33, 2016.

\bibitem{bordes2013translating}
Antoine Bordes, Nicolas Usunier, Alberto Garcia-Duran, Jason Weston, and Oksana
  Yakhnenko.
\newblock Translating embeddings for modeling multi-relational data.
\newblock {\em Advances in neural information processing systems},
  26:2787--2795, 2013.

\bibitem{lin2015learning}
Yankai Lin, Zhiyuan Liu, Maosong Sun, Yang Liu, and Xuan Zhu.
\newblock Learning entity and relation embeddings for knowledge graph
  completion.
\newblock In {\em Twenty-ninth AAAI conference on artificial intelligence},
  pages 2181--2187, 2015.

\bibitem{sun2019rotate}
Zhiqing Sun, Zhi-Hong Deng, Jian-Yun Nie, and Jian Tang.
\newblock Rotate: Knowledge graph embedding by relational rotation in complex
  space.
\newblock {\em arXiv preprint arXiv:1902.10197}, 2019.

\bibitem{nickel2011three}
Maximilian Nickel, Volker Tresp, and Hans-Peter Kriegel.
\newblock A three-way model for collective learning on multi-relational data.
\newblock In {\em Icml}, pages 809--816, 2011.

\bibitem{yang2014embedding}
Bishan Yang, Wen-tau Yih, Xiaodong He, Jianfeng Gao, and Li~Deng.
\newblock Embedding entities and relations for learning and inference in
  knowledge bases.
\newblock {\em arXiv preprint arXiv:1412.6575}, 2014.

\bibitem{nickel2016holographic}
Maximilian Nickel, Lorenzo Rosasco, and Tomaso Poggio.
\newblock Holographic embeddings of knowledge graphs.
\newblock In {\em Proceedings of the AAAI Conference on Artificial
  Intelligence}, volume~30, pages 1955--1961, 2016.

\bibitem{trouillon2016complex}
Th{\'e}o Trouillon, Johannes Welbl, Sebastian Riedel, {\'E}ric Gaussier, and
  Guillaume Bouchard.
\newblock Complex embeddings for simple link prediction.
\newblock In {\em International conference on machine learning}, pages
  2071--2080. PMLR, 2016.

\bibitem{dettmers2018convolutional}
Tim Dettmers, Pasquale Minervini, Pontus Stenetorp, and Sebastian Riedel.
\newblock Convolutional 2d knowledge graph embeddings.
\newblock In {\em Thirty-second AAAI conference on artificial intelligence},
  pages 1811--1818, 2018.

\bibitem{galarraga2013amie}
Luis~Antonio Gal{\'a}rraga, Christina Teflioudi, Katja Hose, and Fabian
  Suchanek.
\newblock Amie: association rule mining under incomplete evidence in
  ontological knowledge bases.
\newblock In {\em Proceedings of the 22nd international conference on World
  Wide Web}, pages 413--422, 2013.

\bibitem{galarraga2015fast}
Luis Gal{\'a}rraga, Christina Teflioudi, Katja Hose, and Fabian~M Suchanek.
\newblock Fast rule mining in ontological knowledge bases with amie++.
\newblock {\em The VLDB Journal}, 24(6):707--730, 2015.

\bibitem{omran2018scalable}
Pouya~Ghiasnezhad Omran, Kewen Wang, and Zhe Wang.
\newblock Scalable rule learning via learning representation.
\newblock In {\em IJCAI}, pages 2149--2155, 2018.

\bibitem{ho2018rule}
Vinh~Thinh Ho, Daria Stepanova, Mohamed~H Gad-Elrab, Evgeny Kharlamov, and
  Gerhard Weikum.
\newblock Rule learning from knowledge graphs guided by embedding models.
\newblock In {\em International Semantic Web Conference}, pages 72--90.
  Springer, 2018.

\bibitem{niu2020rule}
Guanglin Niu, Yongfei Zhang, Bo~Li, Peng Cui, Si~Liu, Jingyang Li, and Xiaowei
  Zhang.
\newblock Rule-guided compositional representation learning on knowledge
  graphs.
\newblock In {\em Proceedings of the AAAI Conference on Artificial
  Intelligence}, volume~34, pages 2950--2958, 2020.

\bibitem{yang2017differentiable}
Fan Yang, Zhilin Yang, and William~W. Cohen.
\newblock Differentiable learning of logical rules for knowledge base
  reasoning.
\newblock In {\em Proceedings of the 31st International Conference on Neural
  Information Processing Systems}, NIPS'17, pages 2316--2325, Red Hook, NY,
  USA, 2017. Curran Associates Inc.

\bibitem{das2017go}
Rajarshi Das, Shehzaad Dhuliawala, Manzil Zaheer, Luke Vilnis, Ishan Durugkar,
  Akshay Krishnamurthy, Alex Smola, and Andrew McCallum.
\newblock Go for a walk and arrive at the answer: Reasoning over paths in
  knowledge bases using reinforcement learning.
\newblock {\em arXiv preprint arXiv:1711.05851}, 2017.

\bibitem{francca2015neural}
Manoel Vitor~Macedo Fran{\c{c}}a, Gerson Zaverucha, and Artur S~d'Avila Garcez.
\newblock Neural relational learning through semi-propositionalization of
  bottom clauses.
\newblock In {\em 2015 AAAI Spring Symposium Series}, 2015.

\bibitem{rocktaschel2017end}
Tim Rocktaschel and Sebastian Riedel.
\newblock End-to-end differentiable proving.
\newblock In {\em Proceedings of the 31st International Conference on Neural
  Information Processing Systems}, NIPS'17, pages 3791--3803, Red Hook, NY,
  USA, 2017. Curran Associates Inc.

\bibitem{minervini2020differentiable}
Pasquale Minervini, Matko Bo{\v{s}}njak, Tim Rockt{\"a}schel, Sebastian Riedel,
  and Edward Grefenstette.
\newblock Differentiable reasoning on large knowledge bases and natural
  language.
\newblock In {\em Proceedings of the AAAI conference on artificial
  intelligence}, volume~34, pages 5182--5190, 2020.

\bibitem{kartavsev2021improving}
Mart Karta{\v{s}}ev, Justin Saler, and Petter {\"O}gren.
\newblock Improving the performance of backward chained behavior trees using
  reinforcement learning.
\newblock {\em arXiv preprint arXiv:2112.13744}, 2021.

\bibitem{xian2020neural}
Yikun Xian, Zuohui Fu, Qiaoying Huang, Shan Muthukrishnan, and Yongfeng Zhang.
\newblock Neural-symbolic reasoning over knowledge graph for multi-stage
  explainable recommendation.
\newblock {\em arXiv preprint arXiv:2007.13207}, 2020.

\bibitem{xian2020cafe}
Yikun Xian, Zuohui Fu, Handong Zhao, Yingqiang Ge, Xu~Chen, Qiaoying Huang,
  Shijie Geng, Zhou Qin, Gerard De~Melo, Shan Muthukrishnan, et~al.
\newblock Cafe: Coarse-to-fine neural symbolic reasoning for explainable
  recommendation.
\newblock In {\em Proceedings of the 29th ACM International Conference on
  Information \& Knowledge Management}, pages 1645--1654, 2020.

\bibitem{minervini2020learning}
Pasquale Minervini, Sebastian Riedel, Pontus Stenetorp, Edward Grefenstette,
  and Tim Rocktschel.
\newblock Learning reasoning strategies in end-to-end differentiable proving.
\newblock In {\em International Conference on Machine Learning}, pages
  6938--6949. PMLR, 2020.

\bibitem{wu2022neural}
Yu-hao Wu and Hou-biao Li.
\newblock Neural theorem provers delineating search area using rnn.
\newblock {\em arXiv preprint arXiv:2203.06985}, 2022.

\bibitem{bodenreider2004unified}
Olivier Bodenreider.
\newblock The unified medical language system (umls): integrating biomedical
  terminology.
\newblock {\em Nucleic acids research}, 32(suppl\_1):D267--D270, 2004.

\bibitem{kok2007statistical}
Stanley Kok and Pedro Domingos.
\newblock Statistical predicate invention.
\newblock In {\em Proceedings of the 24th international conference on Machine
  learning}, pages 433--440, 2007.

\bibitem{jenatton2012latent}
Rodolphe Jenatton, Nicolas Roux, Antoine Bordes, and Guillaume~R Obozinski.
\newblock A latent factor model for highly multi-relational data.
\newblock {\em Advances in neural information processing systems},
  25:3167--3175, 2012.

\bibitem{miller2016key}
Alexander Miller, Adam Fisch, Jesse Dodge, Amir-Hossein Karimi, Antoine Bordes,
  and Jason Weston.
\newblock Key-value memory networks for directly reading documents.
\newblock In {\em Proceedings of the 2016 Conference on Empirical Methods in
  Natural Language Processing}, pages 1400--1409, Austin, Texas, November 2016.
  Association for Computational Linguistics.

\bibitem{tao2002continuous}
Yufei Tao, Dimitris Papadias, and Qiongmao Shen.
\newblock Continuous nearest neighbor search.
\newblock In {\em VLDB'02: Proceedings of the 28th International Conference on
  Very Large Databases}, pages 287--298. Elsevier, 2002.

\bibitem{bouchard2015approximate}
Guillaume Bouchard, Sameer Singh, and Theo Trouillon.
\newblock On approximate reasoning capabilities of low-rank vector spaces.
\newblock In {\em 2015 AAAI Spring Symposium Series}, pages 6--9, 2015.

\end{thebibliography}

\end{document}